# Applying Tabular Deep Learning Models to Estimate Crash Injury Types of Young Motorcyclists

Shriyank Somvanshi
*Ingram School of Engineering, Texas State University*
601 University Drive, San Marcos, Texas 78666
shriyank@txstate.edu

Anannya Ghosh Tusti
*Ingram School of Engineering, Texas State University*
601 University Drive, San Marcos, Texas 78666
gpk30@txstate.edu

Rohit Chakraborty
*Ingram School of Engineering, Texas State University*
601 University Drive, San Marcos, Texas 78666
xuw12@txstate.edu

Subasish Das
*Ingram School of Engineering, Texas State University*
601 University Drive, San Marcos, Texas 78666
subasish@txstate.edu

*Abstract*—Young motorcyclists, particularly those aged 15–24, face a heightened risk of severe crashes due to factors such as speeding, traffic violations, and helmet non-use. This study aims to identify key factors influencing crash severity by analyzing 10,726 young motorcyclist crashes in Texas from 2017 to 2022. Two advanced tabular deep learning models, ARM-Net and MambaNet, were employed, using an advanced resampling technique to address class imbalance. The models were trained to classify crashes into three severity levels: Fatal/Severe (KA), Moderate/Minor (BC), and No Injury (O). ARM-Net achieved an accuracy of 87%, outperforming MambaNet's 86%, with both models excelling in predicting severe and no-injury crashes while facing challenges in moderate crash classification. Key findings highlight the significant influence of demographic, environmental, and behavioral factors on crash outcomes. The study underscores the need for targeted interventions, including stricter helmet enforcement and educational programs customized to young motorcyclists. These insights provide valuable guidance for policymakers in developing evidence-based strategies to enhance motorcyclist safety and reduce crash severity.

## I. INTRODUCTION

Young motorcyclists (ages 15–24) are highly vulnerable to crashes and fatalities due to risky behaviors, limited experience, and factors like speeding and traffic violations [1], [2]. Between 2021 and 2022, fatalities among riders aged 21–24 increased by 16%, with 51% involving speeding [3]. Globally, sensation-seeking, peer influence, and poor helmet use further heighten risks [4]. Rural road conditions and delayed emergency responses exacerbate injury severity [1]. Younger riders often engage in aggressive behaviors and underestimate risks [5], while older riders face more severe injuries due to health factors [6], [7]. Existing safety strategies, such as helmet laws and education campaigns, often overlook age-specific risks.

This study investigates the factors contributing to crashes involving young motorcyclists and aims to identify patterns of influential factors affecting crash injury outcomes. To address gaps in understanding the factors influencing crash severity, the study utilizes Adaptive Relation Modeling Network (ARM-Net) and MambaNet, supported by the Synthetic Minority Oversampling Technique with Edited Nearest Neighbors (SMOTEENN) resampling technique to manage data imbalance. By using these models, the study highlights key risk factors such as road conditions, demographics, environmental factors and offers practical recommendations to enhance infrastructure, safety policies, and overall safety for young motorcyclists.

## II. LITERATURE REVIEW

The comprehensive literature review for this study is organized into two main sections. The first section provides an overview of previous research on safety issues associated with young motorcyclists, while the second section focuses on studies exploring tabular deep learning and its applications.

### A. Studies on Safety Issues associated with Young Motorcyclists

This review section comprehensively examines the factors associated with young motorcyclist crashes, focusing on crash severity and frequency and highlighting the disproportionate involvement of young motorcyclists in severe crashes. Studies report that motorcyclists aged 18–29 have the highest fatality rates in the United States, with those aged 18–20 identified as particularly vulnerable due to speed and rural road use [1], [8]. Studies highlight the need for consistency in age classifications, focusing on riders under 24 and those under 17 [9], [10].

Young motorcyclists face significant traffic safety risks due to behavioral, psychological, and external influences. Key risk factors include traffic and control errors, stunts, protective equipment use, and violations, all strongly linked to crash involvement [11]. Personality traits such as sensation-seeking, impatience, and amiability further contribute, with overconfidence and lack of awareness increasing crash risk [12]. Peer influence also plays a crucial role, especially at unsignalized intersections [13], while psychological factors like perceived behavioral control and susceptibility affect riding behavior [14]. Additionally, delayed reaction times, particularly during early mornings or after sleep deprivation, further elevate crash likelihood [15].

Demographic factors and behavioral norms significantly influence traffic violations, with younger riders being more prone to infractions [16], [17]. Studies have examined how personality traits influence risky behavior across rider groups, while helmet adherence has been shown to vary regionally, with rural motorcyclists demonstrating lower compliance with traffic signs compared to urban riders [12], [18], [19]. Fatal crashes are more common in rural areas than urban regions, with speed and impairment as key contributors [20]. Riders aged 13–19 are significantly more likely to sustain severe injuries than older youth [21]. These findings highlight the importance of understanding regional disparities and using advanced modeling techniques for prediction of fatal crashes to mitigate crash risks for young motorcyclists. In previous study, advanced analytical methods, such as the Bayesian network classifier, improved the prediction of fatal crashes and identified critical risk factors and causal relationships [22].

Theoretical frameworks and evidence-based interventions are critical for addressing risky behaviors among young motorcyclists, with policies focusing on diverse rider profiles proving effective in improving safety outcomes. Studies have identified age-specific challenges, highlighting that motorcyclists under 25 often exhibit poorer safety practices, necessitating strategies such as raising the minimum licensing age to mitigate risks [23]. Normative beliefs, perceived autonomy, and attitudes have been shown to influence risky behaviors, including red-light running and wrong-way driving, emphasizing the importance of behavioral control interventions [17], [24]. These insights underline the need for targeted approaches and advanced crash prediction models to address the diverse behavioral and demographic factors contributing to crash risks among young motorcyclists.

### B. Studies on Tabular Deep Learning

Tabular deep learning models have been widely applied in traffic safety analysis. The evolution of these models, from fully connected networks (FCNs) to advanced architectures like TabNet, SAINT, TabTransformer, and hybrid models, has introduced innovative techniques such as attention mechanisms, feature embeddings, and hybrid approaches to address the complexities of tabular data. These advancements emphasize interpretability, scalability, and efficiency, making them well-suited for traffic safety research [25]. Several studies have specifically explored pedestrian crash severity using the TabNet approach [26], [27]. For example, five deep learning models, including TabNet, were trained and evaluated to predict crash severity in one study [28]. Another study utilized Feature Group Tabular Transformer models for traffic crash modeling and causality analysis [29]. the current study employs ARM-Net [30] and MambaNet [31] to analyze crash severity and uncover patterns in child bicyclists' crashes, providing a deeper understanding of the contributing factors.

### C. Research Gap and Study Contribution

Recent studies on motorcyclist crashes primarily focus on global trends, with limited attention to young motorcyclists in the United States. Critical factors such as risky behaviors, psychological determinants, rule compliance, helmet use, and environmental and demographic influences are often overlooked. Additionally, targeted strategies to address the unique risks faced by young motorcyclists are lacking. This research aims to fill these gaps by analyzing crash data from 2017–2022 in the U.S., exploring key factors affecting crash severity and frequency, and identifying patterns using ARM-Net and MambaNet. The findings will guide the development of effective strategies and countermeasures to reduce young motorcyclist crashes and enhance traffic safety.

## III. DATA PREPARATION

The dataset includes 10,726 young motorcyclist crashes in Texas, categorized into three severity levels: Fatal/Severe (KA), Moderate/Minor (BC), and No Injury (O). Of these, 2,858 crashes were classified as KA, 6,246 as BC (the majority), and 1,622 as O. This stratification enables a detailed examination of crash patterns and the factors contributing to each severity level. Two variable importance analyses identified the important variables that have been considered in this analysis. The variable importance plots, presented in **Figure. 1**, highlight the importance scores using darker shades for higher values. Among the top 18 variables identified, 10 consistently ranked within the top 12 across both models and were selected for the subsequent tabular deep learning analysis. These variables include Road Alignment, Traffic Control Device, First Harmful Event, Other Factors, Road Class, Population Group, Crash Speed Limit, Crash Hour, Person Ethnicity, Person Helmet. The variable codes are presented in TABLE I along with the full variable names.

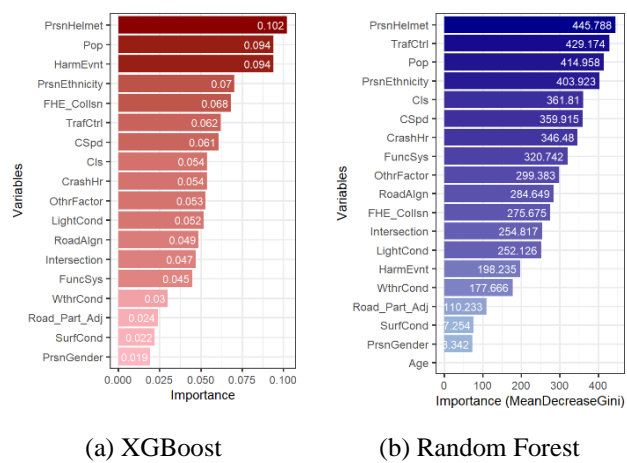

(a) XGBoost      (b) Random Forest

**Figure. 1** Variable importance plot using XGBoost and Random Forest Model

TABLE I lists the distributions of attributes in young motorcyclist crashes across three severity levels. Most crashes occur on "Straight, level" roads, suggesting risks from higher speeds on simpler alignments. "Marked lanes" and "Signal lights" are common crash locations, indicating a need for better intersection and lane management. Crashes are more frequent in urban areas with populations over 250,000 and speed limits of 30–40 mph, reflecting risks from urban congestion and moderate speeds. Additionally, many severe crashes involve riders not wearing helmets or using damaged helmets.

TABLE I DESCRIPTIVE STATISTICS

| Variable name | KA | BC | O |
|---|---|---|---|
|  | N=2858 | N=6246 | N=1622 |
| *Road Alignment (RoadAlgn)* | | | |
| Curve, grade | 154 (5.39%) | 264 (4.23%) | 56 (3.45%) |
| Curve, level | 382 (13.4%) | 624 (9.99%) | 113 (6.97%) |
| Straight, grade | 267 (9.34%) | 453 (7.25%) | 101 (6.23%) |
| Straight, level | 1931 (67.6%) | 4688 (75.1%) | 1292 (79.7%) |
| Other | 124 (4.34%) | 217 (3.47%) | 60 (3.70%) |
| *Traffic Control Device (TrafCtrl)* | | | |
| Marked lanes | 1281 (44.8%) | 2876 (46.0%) | 640 (39.5%) |
| Signal light | 310 (10.8%) | 703 (11.3%) | 258 (15.9%) |
| Stop sign | 297 (10.4%) | 604 (9.67%) | 178 (11.0%) |
| None | 494 (17.3%) | 1251 (20.0%) | 316 (19.5%) |
| Other | 476 (16.7%) | 812 (13.0%) | 230 (14.2%) |
| *First Harmful Event (FHE_Collsn)* | | | |
| Angle-going straight | 273 (9.55%) | 441 (7.06%) | 160 (9.86%) |
| Omv going straight | 1090 (38.1%) | 2642 (42.3%) | 402 (24.8%) |
| One straight-one left | 448 (15.7%) | 596 (9.54%) | 130 (8.01%) |
| rear end | 301 (10.5%) | 682 (10.9%) | 236 (14.5%) |
| Other | 746 (26.1%) | 1885 (30.2%) | 694 (42.8%) |
| *Other Factors (OthrFactor)* | | | |
| Attention diverted from driving | 187 (6.54%) | 481 (7.70%) | 132 (8.14%) |
| One vehicle leaving driveway | 176 (6.16%) | 320 (5.12%) | 110 (6.78%) |
| Vehicle changing lanes | 132 (4.62%) | 433 (6.93%) | 89 (5.49%) |
| Not applicable | 1643 (57.5%) | 2975 (47.6%) | 673 (41.5%) |
| Other | 720 (25.2%) | 2037 (32.6%) | 618 (38.1%) |
| *Road Class (Cls)* | | | |
| City street | 1072 (37.5%) | 2602 (41.7%) | 647 (39.9%) |
| Farm to market | 421 (14.7%) | 740 (11.8%) | 172 (10.6%) |
| Interstate | 355 (12.4%) | 807 (12.9%) | 203 (12.5%) |
| Us&stateHWy | 689 (24.1%) | 1494 (23.9%) | 386 (23.8%) |
| Other | 321 (11.2%) | 603 (9.65%) | 214 (13.2%) |
| *Population Group (Pop)* | | | |
| 100,000-249,999 | 388 (13.6%) | 990 (15.9%) | 217 (13.4%) |
| 250,000&over | 889 (31.1%) | 2409 (38.6%) | 527 (32.5%) |
| 50,000-99,999 | 248 (8.68%) | 568 (9.09%) | 156 (9.62%) |
| Rural | 775 (27.1%) | 1228 (19.7%) | 356 (21.9%) |
| Other | 558 (19.5%) | 1051 (16.8%) | 366 (22.6%) |
| *Crash Speed Limit (CSpd)* | | | |
| 30-40mph | 1148 (40.2%) | 2790 (44.7%) | 783 (48.3%) |
| 45-60mph | 1172 (41.0%) | 2280 (36.5%) | 566 (34.9%) |
| 65-70mph | 301 (10.5%) | 630 (10.1%) | 106 (6.54%) |
| Other | 139 (4.86%) | 255 (4.08%) | 83 (5.12%) |
| Unknown | 98 (3.43%) | 291 (4.66%) | 84 (5.18%) |
| *Crash Hour (CrashHr)* | | | |
| 00-06 | 385 (13.5%) | 597 (9.56%) | 134 (8.26%) |
| 07-12 | 469 (16.4%) | 1249 (20.0%) | 327 (20.2%) |
| 13-18 | 1105 (38.7%) | 2784 (44.6%) | 717 (44.2%) |
| 19-24 | 897 (31.4%) | 1608 (25.7%) | 442 (27.3%) |
| Other | 2 (0.07%) | 8 (0.13%) | 2 (0.12%) |
| *Person Ethnicity (PrsnEthnicity)* | | | |
| Asian | 72 (2.52%) | 161 (2.58%) | 51 (3.14%) |
| Black | 344 (12.0%) | 729 (11.7%) | 191 (11.8%) |
| Hispanic | 719 (25.2%) | 1756 (28.1%) | 395 (24.4%) |
| White | 1643 (57.5%) | 3419 (54.7%) | 900 (55.5%) |
| Other | 80 (2.80%) | 181 (2.90%) | 85 (5.24%) |
| *Person Helmet (PrsnHelmet)* | | | |
| Not worn | 906 (31.7%) | 1677 (26.8%) | 400 (24.7%) |
| Worn, damaged | 1224 (42.8%) | 1989 (31.8%) | 224 (13.8%) |
| Worn, not damaged | 318 (11.1%) | 1502 (24.0%) | 722 (44.5%) |
| Worn, unk damage | 285 (9.97%) | 792 (12.7%) | 151 (9.31%) |
| Other | 125 (4.37%) | 286 (4.58%) | 125 (7.71%) |

*OMV= One Motor Vehicle

## IV. METHODOLOGY

### A. Study Design

This study employs a two-stage methodology to predict crash severity using ARM-Net and MambaNet. ARM-Net leverages an exponential feature transformation and a multi-head gated attention mechanism to capture complex feature interactions while filtering irrelevant attributes, enhancing both prediction accuracy and interpretability [32]. In contrast, MambaNet, a hybrid CNN-LSTM model, excels in capturing spatial and temporal dependencies, effectively modeling relationships between environmental conditions, road characteristics, and driver behaviors [33]. By combining CNNs' spatial feature extraction with LSTMs' sequential learning, MambaNet enhances predictive performance for complex crash severity scenarios, ensuring both short-term and long-term crash patterns are considered.

As illustrated in **Figure 2** in the first stage, the raw crash dataset undergoes preprocessing, beginning with feature selection based on variable importance techniques using Random Forest and XGBoost. The most significant variables identified by these models are analyzed, and the top 10 common features are selected, ensuring relevance to crash severity prediction. To address the issue of class imbalance, the SMOTEENN is applied. This technique combines oversampling and cleaning strategies to balance the dataset by generating synthetic samples for minority classes while simultaneously removing noisy data. Additionally, class weights are computed to further refine the data distribution, resulting in a resampled dataset ready for model training.

In the second stage, the resampled data is used for crash severity prediction. The dataset is split into 60% for training, 20% for validation, and 20% for testing. The deep learning models, ARM-Net and MambaNet, are trained to classify crash severity into three categories: Fatal, Injury, and No Injury. The performance of these models is evaluated using key metrics, including accuracy, precision, recall, F1-score, and confusion matrix analysis. These evaluation measures provide a comprehensive assessment of the model's ability to predict crash severity accurately and reliably.

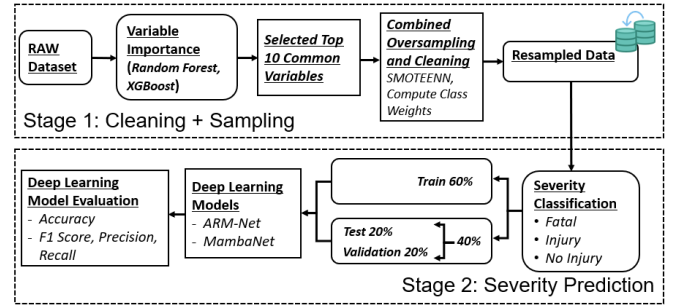

Figure 2 Flowchart of the Combined Study

To mitigate the challenge of imbalanced data, SMOTEENN is applied. SMOTEENN, a hybrid resampling technique combining SMOTE and Edited Nearest Neighbors (ENN), is applied to address data imbalance by oversampling minority classes and removing noisy samples. The resampled dataset includes a total of 8,979 crashes, comprising 3,304 fatal crashes, 1,039 injury crashes, and 4,636 non-injury crashes, ensuring a more balanced distribution. **Figure 3** (feature distribution analysis) confirms that resampling preserves the original data characteristics. The first visualization **Figure 3(a)** shows the overall feature distribution for "Road Alignment, straight" indicating minimal changes post-resampling. The second visualization **Figure 3(b)** presents class-conditional distributions for "Road Alignment, straight", which demonstrates improved class balance while maintaining data integrity. These results validate the effectiveness of SMOTEENN in enhancing crash severity prediction.

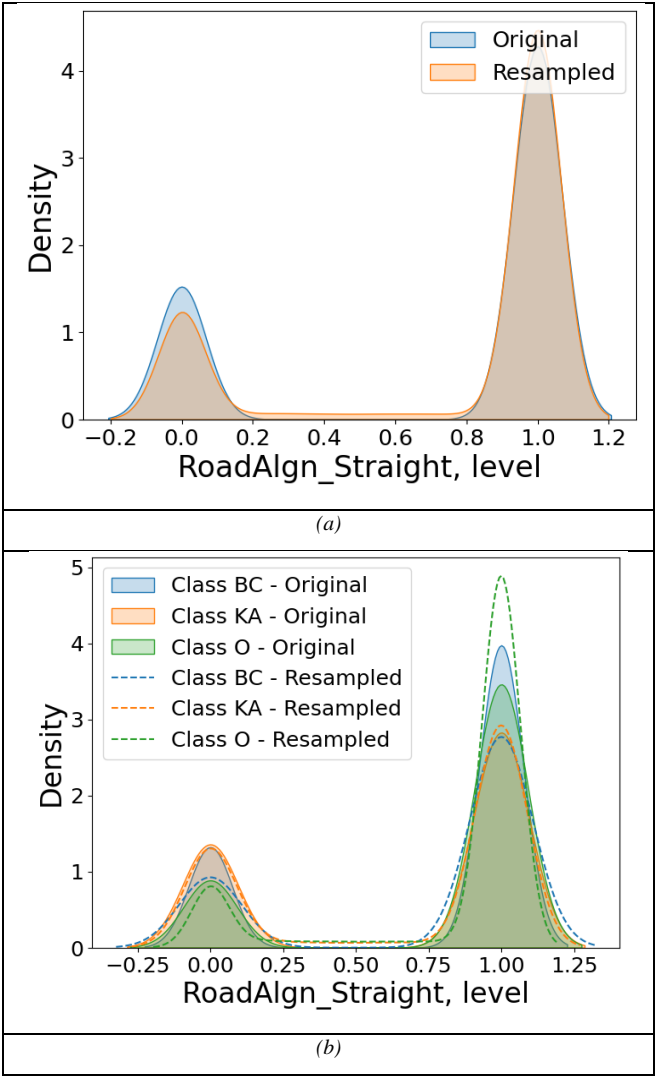

Figure 3 Before and after feature distribution analysis with SMOTEENN (a)Feature distribution of road alignment straight (b) Class-conditional feature distributions of road alignment straight

### B. Hyperparameter Tuning

Hyperparameter optimization is critical for improving the performance of deep learning models such as ARMNet and MambaNet, as their effectiveness depends on selecting optimal parameters. A random search strategy is used to explore the hyperparameter space, sampling 100 configurations per model and training each for 50 epochs. GPU acceleration is utilized to handle computational demands efficiently. ARMNet, designed for tabular data processing, uses attention mechanisms and residual connections with key hyperparameters, including hidden dimensions of [128], four layers, a dropout rate of 0.3, and a batch size of 32. MambaNet employs a multi-layered architecture with dropout strategies to enhance generalization, using hidden dimensions of [128, 64], a dropout rate of 0.3, and a weight decay of 1e-4. This systematic tuning approach ensures robust and generalizable models for crash severity prediction.

Table II HYPERPARAMETER TUNING

| Model | Table Column Head |
|---|---|
| ARM-Net | input_dim: X_train.shape[1], hidden_dim: [128], output_dim: len(y_train.unique()), num_layers: [4], dropout_rate: [0.3], lr: [1e-3], weight_decay: [1e-4], epochs: [50], batch_size: [32], optimizer: AdamW, scheduler: ReduceLROnPlateau |
| MambaNet | input_dim: X_train.shape[1], hidden_dims: [128, 64], output_dim: len(label_encoder.classes_), dropout_rate: [0.3], lr: [1e-3], weight_decay: [1e-4], epochs: [50], batch_size: [32], optimizer: AdamW, scheduler: ReduceLROnPlateau |

## V. RESULTS AND DISCUSSION

### A. Validation of Experiment

**Table III** presents a summary of the training and validation performance of the ARM-Net and MambaNet models for crash severity prediction. The table provides details on the overall accuracy of each model, the number of training epochs, and the distribution of crash severity categories, namely KA (Fatal/Severe), BC (Moderate/Minor), and O (No Injury). ARM-Net achieved a slightly higher accuracy of 87% compared to MambaNet's 86%, indicating its marginally better predictive capability. Both models were trained on a resampled dataset containing 3,304 samples for KA crashes, 1,039 for BC crashes, and 4,636 for O crashes. These findings demonstrate the effectiveness of the models in learning from a balanced dataset and their potential in crash severity classification.

Table III SUMMARY OF TRAINING AND VALIDATION FOR CRASH SEVERITY PREDICTION MODELS

| Model | Accuracy (%) | Epochs | Number of Samples | | |
|---|---|---|---|---|---|
| | | | KA | BC | O |
| ARM-Net | 87 | 50 (Early Stopping) | 3304 | 1039 | 4636 |
| MambaNet | 86 | | 3304 | 1039 | 4636 |

### B. Model Performance

**Table IV** presents a comprehensive evaluation of the prediction performance of ARM-Net and MambaNet across different crash severity categories. The performance metrics assessed include precision, recall, F1-score, and accuracy, providing insights into each model's ability to classify the severity levels KA (Fatal/Severe), BC (Moderate/Minor), and O (No Injury). For ARM-Net, the model demonstrated strong performance in the KA category, achieving a precision of 83%, indicating that 83% of predicted severe crashes were correct. It attained a recall of 92%, reflecting its effectiveness in identifying actual severe crashes. The F1-score of 88% suggests a well-balanced trade-off between precision and recall, with an overall accuracy of 92% in this category. However, ARM-Net struggled in the BC category, where precision dropped to 76% and recall to 37%, leading to an F1-score of 50% and an accuracy of 37%, indicating challenges in correctly identifying moderate/minor crashes. In contrast, the model performed well in the O category, with high precision (91%), recall (94%), and an F1-score of 92%, achieving an overall accuracy of 94%.

MambaNet displayed similar trends, with robust results in the KA category, achieving an 81% precision, 92% recall, and an F1-score of 86%, with an accuracy of 92%. The performance in the BC category was also lower, with precision at 76%, recall at 42%, and an F1-score of 54%, resulting in an accuracy of 42%. In the O category, MambaNet demonstrated strong predictive capability, with a precision of 90%, recall of 91%, and an F1-score of 90%, leading to an accuracy of 91%. Overall, ARM-Net showed slightly better performance in the BC category compared to MambaNet, but both models struggled with moderate/minor crash severity classification. The results suggest that ARM-Net and MambaNet are effective in predicting the extreme severity categories (KA and O), but further improvements may be required to enhance their performance in identifying moderate crash cases.

Table IV PREDICTION PERFORMANCE OF ARM-NET AND MAMBANET MODELS

| Model | Category | Precision (%) | Recall (%) | F-1 Score (%) | Accuracy (%) |
|---|---|---|---|---|---|
| ARM-Net | KA | 83 | 92 | 88 | 92 |
|  | BC | 76 | 37 | 50 | 37 |
|  | O | 91 | 94 | 92 | 94 |
| MambaNet | KA | 81 | 92 | 86 | 92 |
|  | BC | 76 | 42 | 54 | 42 |
|  | O | 90 | 91 | 90 | 91 |

**Fig. 4** presents the confusion matrices comparing the classification performance of the ARM-Net and MambaNet models. MambaNet (**Fig. 4b**) outperforms ARM-Net (**Fig. 4a**) in predicting KA and BC crashes, achieving higher true positive rates and reducing misclassification errors. However, ARM-Net performs better in classifying O (No Injury) cases, with fewer misclassifications in this category. This difference likely stems from MambaNet's hybrid architecture (CNNs + LSTMs), which better captures complex interactions in severe crashes, whereas ARM-Net's attention mechanism is more effective for distinguishing no-injury cases. These findings highlight the models' complementary strengths in crash severity prediction.

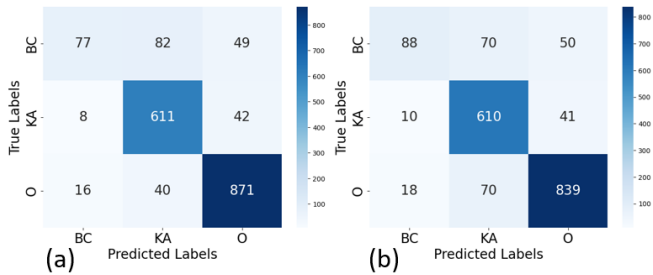

Fig. 4 Confusion Matrix of each model: (a) ARM-Net; (b) MambaNet

The training performance of the ARM-Net and MambaNet models, as shown in the plots, demonstrates effective learning and convergence. For ARM-Net, both training and validation loss (**Fig. 5a**) show a steady decline, indicating consistent learning, while the accuracy curves (**Fig. 5b**) reflect gradual improvement with minimal overfitting. Similarly, MambaNet exhibits a decreasing loss trend with training and validation curves closely aligned, suggesting efficient learning (**Figure 6a**). The accuracy plot for MambaNet (**Figure 6b**) indicates stable improvement over epochs, with validation accuracy surpassing training accuracy at times, highlighting strong generalization capabilities.

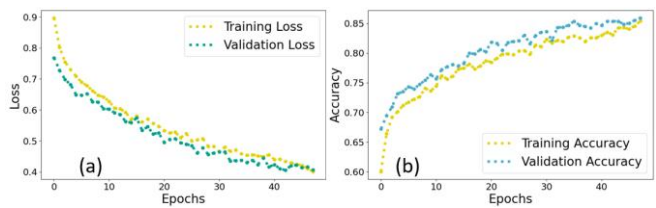

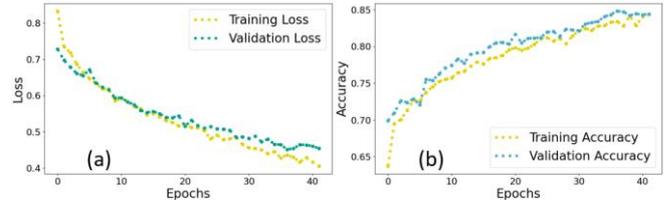

Fig. 5 Training performance of the ARM-Net model: (a) Loss curves; (b) Accuracy curves

Figure 6 Training performance of the MambaNet model: (a) Loss curves; (b) Accuracy curves

## VI. CONCLUSIONS

This study presents a comprehensive analysis of crash severity among young motorcyclists using advanced deep learning models, ARM-Net and MambaNet. By analyzing 10,726 crash records from Texas, the study effectively identifies key factors influencing crash severity and provides valuable insights into patterns that can inform targeted safety interventions. The results indicate that both models exhibit strong predictive capabilities, with ARM-Net achieving a slightly higher accuracy of 87% compared to MambaNet's 86%. Both models struggle to accurately classify moderate injury (BC) crashes, likely due to overlapping feature distributions and biases in crash data. The similarities between crash categories may lead to misclassification, emphasizing the need for more detailed data and improved feature representation. The application of SMOTEENN for data balancing proved effective in mitigating class imbalances and improving model generalization. The study highlights the significance of demographic, environmental, and behavioral factors in influencing crash severity, reinforcing the importance of data-driven approaches in traffic safety analysis. The findings have critical policy implications, emphasizing the need for targeted interventions personalized to the unique risk profiles of young motorcyclists. Stricter enforcement of helmet laws, along with educational programs focused on risk awareness and safe riding practices, can play a crucial role in reducing crash severity. Policymakers should consider integrating AI-driven predictive models into transportation planning to proactively identify high-risk locations and optimize resource allocation. Collaborative efforts between transportation agencies, law enforcement, and public health organizations are essential to implementing comprehensive safety strategies that effectively address young motorcyclists' safety challenges.

Future research can further improve predictive performance by incorporating additional contextual variables such as rider behavior, weather conditions, and traffic patterns. Expanding the dataset to include crash data from diverse regions can enhance model generalizability across different environments. The integration of explainable AI techniques can provide deeper insights into model decision-making, fostering greater trust and usability among stakeholders. Additionally, real-time applications of these models within intelligent transportation systems can support proactive safety measures, enabling authorities to take preventive actions before crashes occur. Integrating ARM-Net and MambaNet into real-time crash prediction systems could enhance traffic safety by leveraging connected vehicle data, sensor-based monitoring, and edge computing for dynamic crash risk assessments. This could enable timely interventions, such as adaptive speed regulation, early hazard warnings, and optimized emergency response. Exploring more advanced deep learning architectures, such as transformer-based models, may also enhance the models' ability to capture complex relationships within crash data.